\documentclass[11pt,letterpaper]{article}
\usepackage{ijcnlp2017}
\usepackage{times}
\usepackage{latexsym}

\usepackage{amsmath,scalerel}
\DeclareMathOperator*{\Bigcdot}{\scalerel*{\cdot}{\bigodot}}
\usepackage{amsfonts}
\usepackage{multirow}
\usepackage{graphicx}
\usepackage{enumerate}
\usepackage{tikz}
\usetikzlibrary{bayesnet}
\usepackage{soul}
\usepackage{linguex}
\usepackage{esvect}
\usepackage{url}
\usepackage{rotating}

\usepackage{color}

\ijcnlpfinalcopy

\title{Distributional Modeling on a Diet: \\One-shot Word Learning from Text Only}

\author{\textbf{Su Wang}$^\spadesuit$\quad\textbf{Stephen Roller}$^\clubsuit$\quad\textbf{Katrin Erk}$^\spadesuit$ \\
 {$^\spadesuit$Department of Linguistics, $^\clubsuit$Department of Computer Science} \\
 {The University of Texas at Austin} \\
 {\tt shrekwang@utexas.edu} \\
 {\tt roller@cs.utexas.edu, katrin.erk@mail.utexas.edu}
 } 
\date{}

\begin{document}

\maketitle

\begin{abstract}
We test whether distributional models can do one-shot learning of
definitional properties 
from text only. Using 
Bayesian models, we find that first learning
overarching structure in the known data, 
regularities in textual contexts and in properties, helps one-shot
learning, and that individual context items can be highly informative.  
Our experiments show that our model can learn properties from a single exposure when given an informative utterance.
\end{abstract}

\section{Introduction}
\label{sec:intro}

When humans encounter an unknown word in text, even with a single instance, they can often
infer approximately what it means, as in this example from
\citet{Lazaridou:2014vt}:
\begin{quote}
We found a cute, hairy \textit{wampimuk} sleeping behind the tree.  
\end{quote}

People who hear this sentence typically guess that a wampimuk
is an animal, or even that it is a mammal. Distributional models,
which describe the meaning of a word in terms of its observed
contexts~\citep{TurneyPantel:10}, have been suggested as a model for
how humans learn word meanings~\citep{Landauer:1997tc}. However,
distributional models typically need hundreds of instances 
of a word to derive a high-quality representation for it, while humans
can often infer a passable meaning approximation from 
one sentence only (as in the above example). This phenomenon is known
as \emph{fast mapping}~\citep{CareyBartlett:78}, 
Our primary modeling objective in this paper is to explore a plausible model for fast-mapping
  learning from textual context.

While there is preliminary evidence that fast mapping can be modeled
distributionally~\citep{Lazaridou:2016tb}, it is unclear what 
enables it. How do humans infer word meanings from so little data?
This question has been studied for \emph{grounded} word learning, when
the learner perceives an object in non-linguistic context that
corresponds to the unknown word. 
The literature emphasizes the importance of learning general knowledge or
overarching structure, which we define as the information that is learned by accumulation across concepts (e.g. regularities in property co-occurrence),
across all concepts~\citep{Kemp:2007}, 
In grounded word learning, overarching structure that has been proposed includes knowledge about which properties. For example knowledge about which properties are most important to object naming ~\citep{Smith:2002,colunga05}, or a taxonomy of
concepts~\citep{XuTenenbaum:07}. 

In this paper we study models for fast mapping in word learning\footnote{In
  this paper, we interchangeably use the terms \emph{unknown word} and
  \emph{unknown concept}, as we learn properties, and properties
  belong to concepts rather than words, and we learn them from text,
  where we observe words rather than concepts.} from textual
context alone, using probabilistic distributional models. Our task differs from the grounded case in that we do not
perceive any object labeled by the unknown word. In that context, learning \emph{word meaning} means learning the associated definitional properties and their weights (see Section 3).
For the sake of interpretability, we focus on 
learning definitional properties
We ask what kinds of overarching structure
in distributional contexts and
in properties will be helpful for one-shot word learning. 


We focus on learning from syntactic context. Distributional
representations of syntactic context are 
directly interpretable as selectional constraints, which in manually
created resources are typically characterized through high-level
taxonomy classes~\citep{VerbNet,FillmoreEtAl:03}. So they should
provide good evidence for the meaning of role fillers. Also, it has
been shown that selectional
constraints can be learned
distributionally~\citep{EPP:Selpref,OSeaghdha:Selpref,Ritter:2010}. 
However, our point will not be that syntax
  is needed for fast word learning, but that it helps to observe
  overarching structure, with syntactic context providing a clear test bed.

We test two types of 
  overarching structure
  for their usefulness in fast mapping. First, we hypothesize that it is
helpful to learn about commonalities among context items, which enables mapping from contexts to properties. For example
the syntactic contexts \textit{eat-dobj} and
\textit{cook-dobj} should prefer similar targets: things that are
 cooked are also things that are eaten (Hypothesis
\textbf{H1}).

The second hypothesis is that it will be
useful to learn co-occurrence patterns between properties. That is, we hypothesize that in learning an entity is a \texttt{mammal}, we may also infer it is \texttt{four-legged} (Hypothesis \textbf{H2}).

We do not intent to make \emph{strong} cognitive claims, for which additional experimentation will be in order, and we leave this for future work. This work sets its goal on building a plausible computational model that models human fast-mapping in learning (i) well from limited grounded data, (ii) effectively from only one instance.



\section{Background}
\label{sec:related}

\textbf{Fast mapping and textual context.} 
Fast mapping~\citep{CareyBartlett:78} is the human ability to
construct provisional word meaning representations after one or few
exposures. An important reason for why humans can do fast mapping is
that they acquire 
  overarching structure that constrains
learning~\citep{Smith:2002,colunga05,Kemp:2007,XuTenenbaum:07,Kemp:09}. 
In this paper, we ask what forms of overarching structure will be useful
for text-based word learning. 

\citet{Lazaridou:2014vt} consider fast mapping for grounded
word learning, mapping image data to distributional representations,
which is in a way the mirror image of our
task. \citet{Lazaridou:2016tb} were the first to explore fast mapping for text-based word learning, using an
extension to word2vec with both textual and visual
features. However, they model the unknown word simply by averaging
the vectors of known words in the sentence, and do not 
explore what types of knowledge enable fast
mapping. 

\textbf{Definitional properties.} Feature norms are
definitional properties collected from human participants. Feature
norm datasets are available from \citet{McRaeEtAlNorms:05}
and \citet{ViglioccoEtAl:04}.  
In this paper we use
  feature norms as our target representations of word meaning.
There are several recent approaches
that learn to map distributional representations to feature
norms~\citep{JohnsJones,rubinstein2015well,Fagarasan:15,Herbelot:2015uz}. We
also map distributional information to 
  feature norms, but we do it based on a single textual instance (one-shot
learning). 

In the current paper we use the \textbf{Quantified McRae (QMR)}
dataset~\citep{Herbelot:15}, which extends the 
\citet{McRaeEtAlNorms:05} feature norms by ratings on 
the proportion of category members that have a property, and the
\textbf{Animal} dataset~\citep{Herbelot:13}, which is smaller but has
the same shape. For example, 
\textit{most} alligators are
dangerous. The quantifiers are given probabilistic interpretations, so
if  \emph{most} alligators are dangerous, the probability for a random
alligator to be dangerous would be 0.95. This makes this dataset a good fit for our
probabilistic distributional model. We discuss QMR and the Animal data
further in Section~\ref{sec:data}. 

\textbf{Bayesian models in lexical semantics.} We use Bayesian models
for the sake of interpretability and because the existing definitional
property datasets are small. The Bayesian models in
lexical semantics that are most related to our approach are \citet{dinu-lapata:2010:EMNLP}, who represent
word meanings as distributions over latent topics that approximate
senses, and \citet{Andrews:15} and \citet{Roller:13}, who use
multi-modal extensions of Latent Dirichlet Allocation (LDA) models~\citep{Blei:03} to represent co-occurrences
of textual context and definitional features. \citet{Seaghdha:2010vi}
  and \citet{Ritter:2010} use Bayesian approaches to model selectional
  preferences.

\section{Models}
\label{sec:model}

\newcommand{\conceptrep}[1]{\ensuremath{\mathbf{#1}}}
\newcommand{\conceptrepInd}[1]{\ensuremath{\mathbf{#1}_{\text{Ind}}}}
\newcommand{\conceptrepMult}[1]{\ensuremath{\mathbf{#1}_{\text{Mult}}}}
\newcommand{\conceptinst}[1]{\ensuremath{\mathbf{\underline{#1}}}}
\newcommand{\alphavec}[1]{\ensuremath{\mathbf{#1}^\alpha}}
\newcommand{\betavec}[1]{\ensuremath{\mathbf{#1}^\beta}}

In this section we develop a series of models to test our hypothesis
that acquiring general knowledge is helpful to word learning, in
particular knowledge about similarities between context items (H1) and
co-occurrences between properties (H2). The count-based model will
implement neither hypothesis, while the bimodal topic model will
implement both. To test the hypotheses separately, we employ two
clustering approaches via Bernoulli Mixtures, which we use as
extensions to the
count-based model and bimodal topic model.

\subsection{The Count-based Model}

\textbf{Independent Bernoulli condition.} Let $Q$ be a set of
definitional properties, $C$ a set of concepts that 
the learner knows about, and $V$ a vocabulary of context
items. For most of our models, context items $w \in V$ will be
predicate-role pairs such as \textit{eat-dobj}. The task is determine
properties that apply to an unknown concept $u \not\in C$. Any concept $c \in C$
is associated with a vector \conceptrepInd{c} (where ``Ind'' stands for
``independent Bernoulli probabilities'') of $|Q|$ probabilities, where the $i$-th
entry of \conceptrepInd{c} is the probability that an instance of concept $c$ would have
property $q_i$. These probabilities are independent Bernoulli
probabilities. For instance, \conceptrepInd{alligator}
would have an entry of 0.95 for \texttt{dangerous}. An \emph{instance}
$\conceptinst{c} \in \{0,1\}^{|Q|}$ 
of a concept $c \in C$ is a vector of zeros and ones drawn from
\conceptrepInd{c}, where an entry of 1 at position $i$ means that this
instance has the property $q_i$. 

The model proceeds in two steps. First it learns property
probabilities for context items $w \in V$. The model observes instances \conceptinst{c}
occurring textually with context item $w$, and learns property
probabilities for $w$, where the
probability that $w$ has for a property $q$ indicates the
probability that $w$ would appear as a context item with an instance
that has property $q$.  In the second step the model uses the acquired
context item representations  to
learn property probabilities for an unknown concept $u$. When $u$
appears with $w$, the context item $w$ ``imagines'' 
an instance (samples it from its property probabilities), and uses this
instance to update the property probabilities of $u$. 
Instead of making point
estimates, the model represents its uncertainty about the probability of a
property through a Beta distribution, a distribution over Bernoulli probabilities. 
As a Beta distribution is characterized by two parameters $\alpha$ and
$\beta$, we associate each context item $w \in V$ with vectors
$\alphavec{w} \in \mathbb{R}^{|Q|}$ and $\betavec{w} \in
  \mathbb{R}^{|Q|}$, where the
  $i$-th $\alpha$ and $\beta$ values are the parameters of the Beta
  distribution for property $q_i$. When an instance \conceptinst{c} is
  observed with context item $w$, we do a Bayesian update on $w$ simply as 
\begin{equation}\label{bayesianupdate}
\begin{array}{l}
\alphavec{w} = \alphavec{w} + \conceptinst{c}\\
\betavec{w} = \betavec{w} + (1 - \conceptinst{c})\\
\end{array}
\end{equation}
because the Beta distribution is the conjugate prior of the
Bernoulli. 
To draw an instance from $w$, we draw it from the predictive posterior
probabilities of its Beta distributions, $\conceptrepInd{w} =
\alphavec{w} / (\alphavec{w} + \betavec{w})$. 

Likewise, we associate an unknown concept $u$ with
vectors $\alphavec{u}$ and $\betavec{u}$. When the model observes $u$
in the context of $w$, it draws an instance from
\conceptrepInd{w}, and performs
a Bayesian update as in (\ref{bayesianupdate}) on the vectors
associated with $u$. After training, the property probabilities for
$u$ are again the posterior predictive probabilities $\conceptrepInd{u} =
\alphavec{u} / (\alphavec{u} + \betavec{u})$. 
The model can be
used for multi-shot learning and one-shot learning in the same way. 

\textbf{Multinomial condition.} We also test a multinomial variant
of the count-based model, for greater comparability with the LDA model
below. Here,  the concept representation
\conceptrepMult{c} is a multinomial distribution over the properties in $Q$. (That
is, all the properties compete in this model.) An instance of concept
$c$ is now a single property, drawn from $c$'s multinomial. The
representation of a context item $w$, and also the representation of
the unknown concept $u$, is a Dirichlet distribution with
$|Q|$ parameters. Bayesian update of the representation of $w$ based
on an occurrence with $c$, and likewise Bayesian update of the
representation of $u$ based on an occurrence with $w$, is straightforward
again, as the Dirichlet distribution is the conjugate prior of the
multinomial.

The two count-based models do not implement either of our two
hypotheses. They compute separate
selectional constraints for each context item, and do not attend
to co-occurrences between properties. In the experiments below, the
count-based models will be listed as 
\textbf{Count Independent} and \textbf{Count Multinomial}.

\subsection{The Bimodal Topic Model}

\begin{figure}[t]
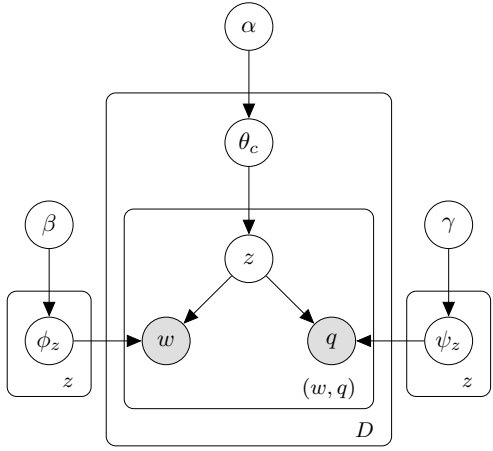

\centering
\resizebox{17em}{34ex}{
\tikz{
	\node[latent](alpha) {$\alpha$} ;
	\node[latent, below=of alpha] (theta) {$\theta_c$} ;
	\node[latent, below=of theta] (z) {$z$} ;
	\node[obs, below left=of z] (w) {$w$} ;
	\node[obs, below right=of z] (q) {$q$} ;
	\node[latent, left=of w] (phi) {$\phi_z$} ;
	\node[latent, right=of q] (psi) {$\psi_z$} ;
	\node[latent, above=of phi] (beta) {$\beta$} ;
	\node[latent, above=of psi] (gamma) {$\gamma$} ;
	\plate[inner sep=0.25cm, xshift=0cm, yshift=0.12cm] {plate1} {(z) (w) (q)} {$(w,q)$}
	\plate[inner sep=0.25cm, xshift=0cm, yshift=0.12cm] {plate2} {(theta) (plate1)} {$D$}
	\plate[inner sep=0.25cm, xshift=0cm, yshift=0.12cm] {plate3} {(phi)} {$z$}
	\plate[inner sep=0.25cm, xshift=0cm, yshift=0.12cm] {plate4} {(psi)} {$z$}
	\edge {alpha} {theta} ;
	\edge {theta} {z} ;
	\edge {z, phi} {w} ;
	\edge {z, psi} {q} ;
	\edge {beta} {phi} ;
	\edge {gamma} {psi} ;
	}
}
      \caption{Plate diagram for the Bimodal Topic Model (bi-TM)}
\label{fig:bimodalTM}
\end{figure}

We use an extension of LDA~\citep{Blei:03} to implement our hypotheses 
on the usefulness of overarching structure, both commonalities in
selectional constraints across predicates, and 
co-occurrence of properties across concepts. In particular, we build on
\citet{Andrews:15} in using a \emph{bimodal topic model}, in which a
single topic simultaneously generates both a context item and a
property. We further build on
\citet{dinu-lapata:2010:EMNLP} in having a 
``pseudo-document'' for each concept $c$ to represent its observed
occurrences. In our case, this pseudo-document contains pairs of a
context item $w \in V$ and a property $q \in Q$, meaning that $w$ has
been observed to occur with an instance of $c$ that had $q$. 

The generative story is as follows. For each known concept $c$, draw
a multinomial $\theta_c$ over topics. For each topic $z$, draw a multinomial $\phi_z$ over context items $w \in V$, and
a multinomial $\psi_z$ over properties $q \in Q$. To generate an entry
for $c$'s pseudo-document, draw a topic $z\sim Mult(\theta_c)$. Then,
from $z$, simultaneously draw a context item from $\phi_z$ and a
property from $\psi_z$. Figure~\ref{fig:bimodalTM} shows the plate
diagram for this model.

To infer properties for an unknown concept $u$, we create a
pseudo-document for $u$ containing just the observed context items, no
properties, as those are not observed. From this pseudo-document $d_u$
we infer the topic distribution
$\theta_u$. Then the probability of a property $q$ given $d_u$ is
\begin{equation}P(q|d_u) = \sum_z P(z | \theta_u)P(q|\psi_z)
\end{equation}
For the one-shot condition, where we only observe a single context
item $w$ with $u$, this simplifies to 
\begin{equation}
P(q|w) = \sum_z P(z|w) P(q|\psi_z)
\end{equation}

We refer to this model as \textbf{bi-TM} below.  The topics of this
model implement our hypothesis H1 by grouping context items that
tend to occur with the same concepts and the same 
properties. The topics also implement our hypothesis H2 by grouping
properties that tend to occur with the same concepts and the same
context items. By using multinomials $\psi_z$ it
makes the simplifying assumption that all properties compete, like the
Count Multinomial model above.

\subsection{Bernoulli Mixtures}

With the Count models, we investigate word learning without any
overarching structures. With the bi-TMs, we investigate word learning
with both types of overarching structures at once. In order to
evaluate each of the two  hypotheses separately, we use clustering
with Bernoulli Mixture models of
either the context items or the properties.


A Bernoulli Mixture model~\citep{JuanVidal:2004} assumes that a
population of $m$-dimensional binary vectors $\mathbf{x}$ has been
generated by a set of mixture components $K$, each of which is a vector of $m$
Bernoulli probabilities:
\begin{equation}
p(\mathbf{x}) = \sum_{k=1}^{|K|} p(k) p(\mathbf{x}|k)  
\end{equation}
A Bernoulli Mixture can represent co-occurrence patterns between the $m$
random variables it models without assuming competition between them. 

To test the effect of modeling \emph{cross-predicate selectional
  constraints}, we estimate a Bernoulli Mixture model from $n$ 
instances \conceptinst{w} for each $w\in V$, sampled from
\conceptrepInd{w} (which is learned as in the Count Independent model).
Given a Bernoulli Mixture model of $|K|$ components, we then assign
each context item $w$ to its closest mixture component as
follows. Say the instances of $w$ used to estimate the Bernoulli Mixture were
$\{\conceptinst{w}_1,\ldots,\conceptinst{w}_n\}$,  then we assign $w$
to the component 
\begin{equation}
k_w = \text{argmax}_k \sum_{j=1}^n p(k|\conceptinst{w}_j)  
\end{equation}
We then re-train the representations of context items in the Count
Multinomial condition, treating each occurrence of $c$ with context
$w$ as an occurrence of $c$ with $k_w$. This yields a Count
Multinomial model called \textbf{Count BernMix H1}.

To test the effect of modeling \emph{property co-occurrences}, we
estimate a $|K|$-component Bernoulli Mixture model from $n$ 
instances of each known concept $c \in C$, sampled from
\conceptrepInd{c}. We then represent each concept $c$ by a vector
\conceptrepMult{c}, a multinomial with $|K|$ parameters, as
follows. Say the instances of $c$ used to estimate the Bernoulli Mixture were
$\{\conceptinst{c}_1,\ldots,\conceptinst{c}_n\}$,  then the $k$-th
entry in \conceptrepMult{c} is the average probability, over all
$\conceptinst{c}_i$, of being generated by  component $k$:
\begin{equation}
\conceptrep{c}_k = \frac{1}{n} \sum_{j=1}^n p(k|\conceptinst{c}_j)  
\end{equation}
This can be used as a Count Multinomial model where the entries in
\conceptrepMult{c} stand for Bernoulli Mixture components rather than
individual properties. We refer to it as \textbf{Count BernMix
  H2}.\footnote{We use the H2 Bernoulli Mixture as a soft
  clustering because it is straightforward to do this
  through concept representations. For
  the H1 mixture, we did not see an obvious soft clustering,
  so we use it as a hard clustering.}

Finally, we extend the bi-TM with the H2 Bernoulli Mixture in the same
way as a Count Multinomial model, and list this extension as
\textbf{bi-TM BernMix H2}. While the bi-TM already implements both H1
and H2, its assumption of competition between all properties is
simplistic, and bi-TM BernMix H2 tests whether lifting this assumption
will yield a better model. We do not extend the bi-TM with the H1
Bernoulli Mixture, as the assumption of competition between context
items that the bi-TM makes is appropriate.

\section{Data and Experimental Setup}
\label{sec:data}

\textbf{Definitional properties.} As we use probabilistic models, we need probabilities of properties applying to concept instances. So the QMR dataset~\citep{Herbelot:15} is ideally suited. QMR has 532 concrete noun concepts, each associated with a set of quantified properties. The quantifiers have been given probabilistic interpretations, mapping 
\texttt{all}$\rightarrow$1, \texttt{most}$\rightarrow$0.95, \texttt{some}$\rightarrow$0.35, \texttt{few}$\rightarrow$0.05, \texttt{none}$\rightarrow$0.\footnote{The dataset also contains \texttt{KIND} properties that do not have probabilistic interpretations. Following \citet{Herbelot:2015uz} we omit these properties.} Each concept/property pair was judged by 3 raters. We choose the majority rating when it exists, and otherwise the minimum proposed rating. To address sparseness, especially for the one-shot learning setting, we omit properties that are named for fewer than 5 concepts. This leaves us with 503 concepts and 220 properties
We intentionally choose this small dataset: One of our main objectives is to explore the possibility of learning effectively from very limited training data. In addition, while the feature norm dataset is small, our distributional dataset (the BNC, see below) is not. The latter essentially serves as a pivot for us to propagate the knowledge from the feature norm data to the wider semantic space.

It is a problem of both the original \citet{McRaeEtAlNorms:05} data and QMR that if a property is not named by participants, it is not listed, even if it applies. For example, the property \texttt{four-legged} is missing for \textit{alligator} in QMR. So we additionally use the \textbf{Animal} dataset of \citet{Herbelot:13}, where every property has a rating for every concept. The dataset comprises 72 animal concepts with quantification information for 54 properties. 

\textbf{Distributional data.} We use the British National Corpus (BNC)~\citep{BNC:07},  with dependency parses from Spacy.~\footnote{\url{https://spacy.io}} As context items, we use pairs $\langle$pred, dep$\rangle$ of predicates pred that are content words (nouns, verbs, adjectives, adverbs) but not stopwords, where a concept from the respective dataset (QMR, Animal) is a dependency child of pred via dep. 
In total we obtain a vocabulary of 500 QMR concepts and 72 Animal concepts that appear in the BNC, and 29,124 context items. We refer to this syntactic context as \textbf{Syn}. For comparison, we also use a baseline model with a bag-of-words (\textbf{BOW}) context window of 2 or 5 words, with stopwords removed.

\textbf{Models.} We test our probabilistic models as defined in the previous section. While our focus is on one-shot learning, we also evaluate a multi-shot setting where we learn from the whole BNC, as a sanity check on our models. (We do not test our models in an incremental learning setting that adds one occurrence at a time. While this is possible in principle, the computational cost is prohibitive for the bi-TM.) We compare to the Partial Least Squares (\textbf{PLS}) model of \citet{Herbelot:2015uz}\footnote{\citet{Herbelot:2015uz} is the only directly relevant previous work on the subject. Further, to the best of our knowledge, for one-shot property learning from text (only), our work has been the first attempt.} to see whether our models perform at state of the art levels. We also compare to a baseline that always predicts the probability of a property to be its relative frequency in the set $C$ of known concepts (\textbf{Baseline}). 

We can directly use the property probabilities in QMR and the Animal data as concept representations \conceptrepInd{c} for the Count Independent model. For the Count Multinomial model, we never explicitly compute 
\conceptrepMult{c}. To sample from it, we first sample an instance $\conceptinst{c} \in \{0,1\}^{|Q|}$ from the independent Bernoulli vector of $c$, \conceptrepInd{c}. From the properties that apply to \conceptinst{c}, we sample one (with equal probabilities) as the observed property. All priors for the count-based models (Beta priors or Dirichlet priors, respectively) are set to 1.

For the bi-TM, a pseudo-document for a known concept $c$ is generated as follows: Given an occurrence of known concept $c$ with context item $w$ in the BNC, we sample a property $q$ from $c$ (in the same way as for the Count Multinomial model), and add $\langle w, q\rangle$ to the pseudo-document for $c$. For training the bi-TM, we use collapsed Gibbs sampling \citep{Steyvers:07} with 500 iterations for burn-in. 
The Dirichlet priors are uniformly set to 0.1 following \citet{Roller:13}. We use 50 topics throughout. 



For all our models, we report the average performance from 5 runs.
For the PLS benchmark, we use 50 components  with otherwise default settings, following \citet{Herbelot:2015uz}.

\textbf{Evaluation.} 
We test all models using 5-fold cross validation and report average performance across the 5 folds. 
We evaluate performance using \textit{Mean Average Precision} (MAP) 
, which tests to what extent a model ranks definitional properties in the same order as the gold data. Assume a system that predicts a ranking of $n$ datapoints, where 1 is the highest-ranked, and assume that each datapoint $i$ has a gold rating of $I(i) \in \{0, 1\}$. This system obtains an Average Precision (AP) of 
\[AP = \frac{1}{\sum_{i=1}^n I(i)} \sum_{i=1}^n \text{Prec}_i \cdot I(i)\]
where $\text{Prec}_i$ is precision at a cutoff of $i$. Mean Average Precision is the mean over multiple AP values. 
In our case, $n = |Q|$, and we compare a model-predicted ranking of property probabilities with a binary gold rating of whether the property applies to any instances of the given concept.
For the one-shot evaluation, we make a separate prediction for each occurrence of an unknown concept $u$ in the BNC, and report MAP by averaging over the AP values for all occurrences of $u$.

\section{Results and Discussion}
\label{sec:results}

\begin{table}[t]
	\small
	\centering
	\begin{tabular}{|ll|cc|c|}
		\hline \multicolumn{2}{|l|}{\multirow{2}{*}{Models}} & \multicolumn{2}{c|}{QMR} & Animal \\
		&& BOW5 & Syn & Syn \\ \hline\hline
		\multicolumn{2}{|l|}{Baseline} & 0.12 & 0.16 & 0.63 \\\hline
		PLS && \textbf{0.24} & 0.35 & 0.71 \\\hline
		Count & Mult. & 0.13 & 0.25 & 0.64 \\
		 & Ind. & 0.11 & 0.23 & 0.64 \\
		 & BernMix H1 & 0.11 & 0.17 & 0.65 \\
		 & BernMix H2 & 0.10 & 0.18 & 0.63 \\\hline
		bi-TM & plain & 0.23 & \textbf{0.36} & 0.80 \\
		& BernMix H2 & 0.20 & 0.34 & \textbf{0.81} \\ \hline
	\end{tabular}
	\caption{MAP scores, multi-shot learning on the QMR and Animal datasets}
        \label{tab:multi-shot-all} 
\end{table}

\textbf{Multi-shot learning.} 
While our focus in this paper
  is on one-shot learning, we first test all models in a multi-shot
  setting. The aim is to see how well they perform when given ample
  amounts of training data, and to be able to compare their performance to an
  existing multi-shot model (as we will not have any related work to
  compare to for the one-shot setting.)
The results are
shown in Table~\ref{tab:multi-shot-all}, where \textit{Syn} shows
results that use syntactic context (encoding selectional constraints)
and \textit{BOW5} is a bag-of-words context with a window size of
5. We only compare our models to the baseline and benchmark for now,
and do an in-depth comparison of our models when we get to the
one-shot task, which is our main focus. 

Across all models, the syntactic context outperforms the
bag-of-words context. We also tested a bag-of-words context with
window size 2 and found
it to have a performance halfway between \textit{Syn} and \textit{BOW5}
throughout. This confirms our assumption that it is reasonable to
focus on syntactic context, and for the rest of this paper, we test
models with syntactic context only. 

Focusing on \textit{Syn} conditions now, we see that almost all models
outperform the property frequency baseline, though the MAP scores for
the baseline do not fall far behind those of the weakest count-based
models.\footnote{This is because MAP gives equal
credit for all properties correctly predicted as non-zero. When we
evaluate 
with Generalized Average Precision (GAP)~\citep{kishida:2005:gap}, which takes gold
 weights into account, the baseline model is roughly 10 points below other models. This indicates our models learn approximate property distributions. We omit GAP scores because they correlate strongly with MAP for non-baseline models.} The best of our models perform on par
with the  PLS benchmark of \citet{Herbelot:2015uz} on QMR, and on the
Animal dataset they outperform the benchmark.
Comparing the two datasets, we see that all models show better
performance on the cleaner (and smaller) \textit{Animal} dataset than on
QMR. This is probably because QMR suffers from many false negatives (properties that
apply but were not mentioned), while Animal does not. The
Count Independent model shows similar performance here and
throughout all later experiments to the Count Multinomial (even
though it matches the construction of the QMR and Animal datasets
better), so to avoid clutter we do not report on it further below.


\begin{table}[t]
	\small
	\centering
	\begin{tabular}{|l|ll|ccc|}
		\hline 
\multicolumn{1}{|l}{~}& \multicolumn{2}{l|}{\multirow{2}{*}{Models}}  && oracle & AvgCos\\
\multicolumn{1}{|l}{~}&& & all & top20 & top20\\\hline\hline
\multirow{5}{*}{\begin{sideways}QMR\end{sideways}} & 
Count & Mult. & 0.16 & 0.37 & 0.28\\
& & BernMix H1 & 0.14 & 0.33 & 0.21\\
& & BernMix H2 & 0.15 & 0.31 & 0.22\\\cline{2-6}
&bi-TM & plain & \textbf{0.21} & \textbf{0.47} & \textbf{0.35}\\
& & BernMix H2 & 0.18 & 0.45 & 0.34\\\hline\hline
\multirow{5}{*}{\begin{sideways}Animal\end{sideways}}
& Count & Mult. & 0.58 & 0.77 & 0.61 \\
& & BernMix H1 & 0.60 & 0.80 & 0.57\\
& & BernMix H2 & 0.59 & 0.81 & 0.59\\\cline{2-6} 
& bi-TM & plain & 0.64 & 0.88 & 0.63\\
&& BernMix H2 & \textbf{0.65} & \textbf{0.89} & \textbf{0.66}\\\hline
	\end{tabular}
	\caption{MAP scores, one-shot learning on the QMR and
          Animal datasets}
\label{tab:one-shot-ap} 
\end{table}



\textbf{One-shot learning.} Table~\ref{tab:one-shot-ap} shows the performance of
our models on the one-shot learning task. We cannot evaluate the
benchmark PLS as it is not suitable for one-shot learning. The
baseline is the same as in Table~\ref{tab:multi-shot-all}. The
numbers shown are Average Precision (AP) values for learning from a single
occurrence. Column \textit{all} averages over all occurrences of a
target in the BNC (using only context items that appeared at least 5
times in the BNC), and column \textit{oracle top-20} averages over the
20 context items that have the highest AP for the given target. As can be
seen, AP varies widely across sentences: When we average over all
occurrences of a target in the BNC, performance is close to baseline
level.\footnote{Context items with few occurrences in the corpus
  perform considerably worse than baseline, as their property
  distributions are dominated by the small number of concepts with
  which they appear.}
But the
most \emph{informative} instances yield excellent information about an
unknown concept, and lead to MAP values that are much higher than those achieved in
multi-shot learning
(Table~\ref{tab:multi-shot-all}). We explore this more below.

Comparing our models, we see that the bi-TM does much better
throughout than any of the count-based models. 
Since the bi-TM model implements both cross-predicate selectional
constraints (H1) and property co-occurrence (H2), we find both of our
hypotheses confirmed by these results. 
The Bernoulli mixtures improved performance on the Animal
dataset, with no clear pattern of which one improved performance
more. On QMR, adding a Bernoulli mixture model harms
performance across both the 
count-based and bi-TM models. We suspect that this is because of the
false negative entries in QMR; an inspection of Bernoulli mixture H2
components supports this intuition, as the QMR ones were found to be of
poorer quality than those for the Animal data. 

Comparing Tables~\ref{tab:multi-shot-all}  and \ref{tab:one-shot-ap}
we see that they show the same patterns of performance: Models that do
better on the multi-shot task also do better on the one-shot
task. This is encouraging in that it suggests that it should be
possible to build incremental models that do well both in a low-data
and an abundant-data setting.

\begin{table}[t]
	\small
	\centering
	\begin{tabular}{|p{3.5em}|p{18em}|}
		\hline
		Count Mult. & \texttt{clothing},
                             \texttt{made\_of\_metal},
                              \texttt{dif\-fer\-ent\_colours}, \texttt{an\_animal}, \texttt{is\_long} \\\hline
		bi-TM & \texttt{clothing},
                         \texttt{made\_of\_material},
                         \texttt{has\_\-sleeves},
                         \texttt{different\_colours},
                         \texttt{worn\_by\_women} \\\hline
bi-TM one-\-shot & 
          \texttt{clothing}, \texttt{is\_long},
                                             \texttt{made\_\-of\_\-material},
                                             \texttt{different\_colours}, 
\texttt{has\_sleeves}\\
		\hline
	\end{tabular}
	\caption{QMR: top 5 properties of \textit{gown}. Top 2
          entries: multi-shot. Last entry: one-shot, context \textit{undo-dobj}}
\label{tab:count-vs-btm-gowndemo} 
\end{table}

Table~\ref{tab:count-vs-btm-gowndemo}  looks in more detail at what it is
that the models are learning by showing the five highest-probability
properties they are predicting for the concept \textit{gown}. The top two entries are multi-shot
models, the third shows the one-shot result from the context item with
the highest AP. The bi-TM results are very good in both the multi-shot
and the one-shot setting, giving high probability to some quite specific
properties like \texttt{has\_sleeves}. The count-based model shows a
clear frequency bias in erroneously giving high probabilities to 
the two overall most frequent properties, \texttt{made\_of\_metal} and
\texttt{an\_animal}. This is due to the additive nature of the
Count model: In updating unknown concepts from context items, frequent
properties are more likely to be sampled, and their effect
accumulates as the model does not take into account interactions among context
 items. The bi-TM, which models these interactions, is much more robust
 to the effect of property frequency.

\begin{table}[t]
	\small
	\centering
	\begin{tabular}{|l|p{17em}|}
		\hline
		Top & \textit{undo-dobj} (0.70), \textit{nylon-nmod} (0.66),
                      \textit{pink-amod} (0.65), \textit{retie-dobj} (0.64),
                      \textit{silk-amod} (0.64) \\\hline
		Bottom  & \textit{sport-nsubj} (0.01),
                          \textit{contemplate-dobj} (0.01), 
                          \textit{comic-amod}  (0.01), \textit{wait-nsubj} (0.01),
                          \textit{fibrous-amod} (0.01) \\\hline
	\end{tabular}
	\caption{QMR one-shot: AP for top and  bottom 5 context items of
          \textit{gown}}
\label{tab:one-shot-gowndemo} 
\end{table}

 \textbf{Informativity.}
In Table~\ref{tab:one-shot-ap} we saw that one-shot performance
averaged over all context items in the whole corpus was quite bad, but
that good, \emph{informative} context items can yield high-quality
property information. Table~\ref{tab:one-shot-gowndemo} illustrates
this point further. For the concept \textit{gown}, it shows the five
context items that yielded the highest AP values, at the top
\textit{undo-obj}, with an AP as high as 0.7.

\begin{table}[t]
	\small
	\centering
	\begin{tabular}{|l|ll|lll|}\hline
		\multicolumn{1}{|l}{~}& \multicolumn{2}{|l}{Model} & Freq.& Entropy & AvgCos
          \\ \hline
\multirow{5}{*}{\begin{sideways}QMR\end{sideways}} & 
Count & Mult. & 0.09 & -0.12 & 0.18 \\
& Count & BernMix H1 & 0.07 & -0.10 & 0.17 \\
& Count &  BernMix H2 & 0.10 & -0.09 & 0.17 \\
& bi-TM  & plain & 0.15 & -0.09 & 0.41$^{\Bigcdot}$ \\
& bi-TM & BernMix H2 & 0.16 & -0.10 & 0.39$^{\Bigcdot}$ \\ \hline
\multirow{2}{*}{\begin{sideways}Ani.\end{sideways}} & 
bi-TM  & plain & 0.25 & -0.40 & 0.49* \\
& bi-TM & BernMix H2 & 0.23$^{\Bigcdot}$ & -0.37$^{\Bigcdot}$ & 0.52* \\ \hline
	\end{tabular}
	\caption{Correlation of informativity with AP, Spearman's
          $\rho$. * and $^{\Bigcdot}$ indicate significance at $p<0.05$ and $p<0.1$}
       \label{tab:one-shot-informativity} 
\end{table}
 
This raises the question of whether we can predict
the informativity of a context item.\footnote{\citet{Lazaridou:2016tb}, who use
a bag-of-words context in one-shot experiments, propose
an informativity measure based on the number of contexst that constitute properties. we cannot do that with our syntactic context.} We test
three measures of informativity. The first is simply the 
\textbf{frequency} of the context item, with the rationale that more frequent
context items should have more stable representations. Our second measure is based on \textbf{entropy}. For
each context item $w$, we compute a distribution over properties as in
the count-independent model, and measure the entropy of this
distribution. If the distribution has few properties account for 
a majority of the probability mass, then $w$ will have a low entropy,
and would be expected to be more informative. 
Our third measure is
based on the same intuition, that items with more ``concentrated''
selectional constraints should be more informative. If a context item
$w$ has been observed to occur with known concepts $c_1, \ldots, c_n$,
then this measure is the average cosine (\textbf{AvgCos}) of the
property distributions (viewed as vectors) of any pair of $c_i, c_j \in \{c_1,
\ldots, c_n\}$. 

We evaluate the three informativity measures using Spearman's rho to
determine the correlation of the informativity of a context item with the AP it
produces for each unknown concept. We expect frequency and AvgCos to be
positively correlated with AP, and entropy to be negatively
correlated with AP. The result is shown in
Table~\ref{tab:one-shot-informativity}.  Again, all measures work better on
the Animal data than on QMR, where they at best approach
significance. The correlation is much better on the bi-TM models
than on the count-based models, which is probably due to their higher-quality
predictions. Overall, AvgCos emerges as the most robust indicator for 
informativity.\footnote{We also tested a
  binned variant of the frequency measure, on the intuition that
  medium-frequency context items should be more informative than
  either highly frequent or rare ones. However, this measure did not
  show better performance than the non-binned frequency measure.} 
We now test AvgCos, as our best informativity measure, on its ability to
select good context items. The last column of
Table~\ref{tab:one-shot-ap} shows MAP results for the top 20
context items based on their AvgCos values. The results
are much below the oracle MAP (unsurprisingly, given the correlations
in Table~\ref{tab:one-shot-informativity}), but for QMR they are at
the level of the multi-shot results of Table~\ref{tab:multi-shot-all},
showing that it is possible to some extent to automatically choose
informative examples for one-shot learning.


\begin{table}
	\small
	\centering
\begin{tabular}{|l|l|}\hline
Type & MAP\\\hline\hline
Function & 0.45\\
Taxonomic & \textbf{0.62}\\
Visual & 0.34\\
Encyclopaedic & 0.35\\
Perc & 0.40\\\hline
	\end{tabular}
	\caption{QMR, bi-TM, one-shot: MAP by property type over (oracle) top 20
          context items}	
\label{tab:ap-by-property-type} 
\end{table}

\textbf{Properties by type.}
\citet{McRaeEtAlNorms:05} classify properties based on the brain
region taxonomy of \citet{CreeMcRae:03}. This enables us to test what
types of properties are learned most easily in our fast-mapping
setup by computing average AP separately by property type. To combat
sparseness, we group property types into five groups,
\textit{function} (the function or use of an entity),
\textit{taxonomic}, \textit{visual}, \textit{encyclopaedic},  and
\textit{other perceptual} (e.g., sound). Intuitively, we would expect
our contexts to best reflect \textit{taxonomic} and \textit{function}
properties: Predicates that apply to noun target concepts often
express functions of those targets, and manually specified selectional constraints are
often characterized in terms of taxonomic classes. 
Table~\ref{tab:ap-by-property-type} confirms this intuition. Taxonomic properties achieve
the highest MAP by a large margin, followed by functional
properties. Visual properties score the lowest.

\section{Conclusion}
\label{sec:conclusion}

We have developed several models for one-shot learning word meanings from single textual contexts. Our models were designed learn word properties using distributional contexts (H1) or about co-occurrences of properties (H2). We find evidence that both kinds of general knowledge
are helpful, especially when combined (in the bi-TM), or when used on clean property data (in the Animal dataset). We further saw that some contexts are highly informative, and preliminary expirements in informativity measures found that average pairwise similarity of seen role fillers (AvgCos) achieves some success in predicting which contexts are most useful.

In the future, we hope to test with other types of general knowledge, including a taxonomy of known concepts~\citep{XuTenenbaum:07}; wider-coverage property data \citep[Type-DM]{Baroni:10}; and alternative modalities \citep[image features as ``properties'']{Lazaridou:2016tb}. We expect our model will scale to these larger problems easily.

We would also like to explore better informativity measures and improvements for AvgCos.
Knowledge about informative examples can be useful in human-in-the-loop settings, for example a user aiming to illustrate classes in an ontology with a few typical corpus examples. 
We also note that the bi-TM cannot be used in for truly incremental learning, as the cost of global re-computation after each seen example is prohibitive. We would like to explore probabilistic models that support incremental word learning, which would be interesting to integrate with an overall probabilistic model of semantics~\citep{GoodmanLassiter}.

\section*{Acknowledgments}

This research was supported by the DARPA DEFT program under AFRL grant FA8750-13-2-0026 and
by the NSF CAREER grant IIS 0845925. Any opinions, findings, and conclusions or recommendations
expressed in this material are those of the author and do not necessarily reflect the view of DARPA, DoD
or the US government.  We acknowledge the Texas Advanced Computing Center for
providing grid resources that contributed to these results.

\bibliography{ijcnlp2017}

\begin{thebibliography}{35}
\expandafter\ifx\csname natexlab\endcsname\relax\def\natexlab#1{#1}\fi

\bibitem[{Andrews et~al.(2009)Andrews, Vigliocco, and Vinson}]{Andrews:15}
Mark Andrews, Gabriella Vigliocco, and David Vinson. 2009.
\newblock Integrating experiential and distributional data to learn semantic
  representations.
\newblock \emph{Psychological Review}, 116(3):463--498.

\bibitem[{Baroni and Lenci(2010)}]{Baroni:10}
Marco Baroni and Alexandero Lenci. 2010.
\newblock Distributional memory: a general framework for corpus-based
  semantics.
\newblock \emph{Computational Linguistics}, 36(4):673--721.

\bibitem[{Blei et~al.(2003)Blei, Ng, and Jordan}]{Blei:03}
David Blei, Andrew Ng, and Michael Jordan. 2003.
\newblock Latent {Dirichlet} {Allocation}.
\newblock \emph{Journal of Machine Learning Research}, 3(4-5):993--1022.

\bibitem[{Carey and Bartlett(1978)}]{CareyBartlett:78}
Susan Carey and Elsa Bartlett. 1978.
\newblock Acquiring a single new word.
\newblock \emph{Papers and Reports on Child Language Development}, 15:17--29.

\bibitem[{Colunga and Smith(2005)}]{colunga05}
Eliana Colunga and Linda~B. Smith. 2005.
\newblock From the lexicon to expectations about kinds: A role for associative
  learning.
\newblock \emph{Psychological Review}, 112(2):347--382.

\bibitem[{Cree and McRae(2003)}]{CreeMcRae:03}
George~S. Cree and Ken McRae. 2003.
\newblock Analyzing the factors underlying the structure and computation of the
  meaning of chipmunk, cherry, chisel, cheese, and cello (and many other such
  concrete nouns).
\newblock \emph{Journal of Experimental Psychology: General}, 132:163--201.

\bibitem[{Dinu and Lapata(2010)}]{dinu-lapata:2010:EMNLP}
Georgiana Dinu and Mirella Lapata. 2010.
\newblock Measuring distributional similarity in context.
\newblock In \emph{Proceedings of EMNLP}, Cambridge, MA.

\bibitem[{Erk et~al.(2010)Erk, Pad\'{o}, and Pad\'{o}}]{EPP:Selpref}
Katrin Erk, Sebastian Pad\'{o}, and Ulrike Pad\'{o}. 2010.
\newblock A flexible, corpus-driven model of regular and inverse selectional
  preferences.
\newblock \emph{Computational Linguistics}, 36(4).

\bibitem[{F{\u a}g{\u a}r{\u a}{\c s}an et~al.(2015)F{\u a}g{\u a}r{\u a}{\c
  s}an, Vecchi, and Clark}]{Fagarasan:15}
Luana F{\u a}g{\u a}r{\u a}{\c s}an, Eva~Maria Vecchi, and Stephen Clark. 2015.
\newblock From distributional semantics to feature norms: Grounding semantic
  models in human perceptual data.
\newblock In \emph{Proceedings of IWCS}, London, Great Britain.

\bibitem[{Fillmore et~al.(2003)Fillmore, Johnson, and
  Petruck}]{FillmoreEtAl:03}
C.~J. Fillmore, C.~R. Johnson, and M.~Petruck. 2003.
\newblock Background to {F}rame{N}et.
\newblock \emph{International Journal of Lexicography}, 16:235--250.

\bibitem[{Goodman and Lassiter(2014)}]{GoodmanLassiter}
Noah~D. Goodman and Daniel Lassiter. 2014.
\newblock Probabilistic semantics and pragmatics: Uncertainty in language and
  thought.
\newblock In Shalom Lappin and Chris Fox, editors, \emph{Handbook of
  Contemporary Semantics}. Wiley-Blackwell.

\bibitem[{Herbelot(2013)}]{Herbelot:13}
Aur\'{e}lie Herbelot. 2013.
\newblock What is in a text, what isn't and what this has to do with lexical
  semantics.
\newblock \emph{Proceedings of IWCS}.

\bibitem[{Herbelot and Vecchi(2015{\natexlab{a}})}]{Herbelot:2015uz}
Aur\'{e}lie Herbelot and Eva Vecchi. 2015{\natexlab{a}}.
\newblock Building a shared world:mapping distributional to model-theoretic
  semantic spaces.
\newblock In \emph{Proceedings of EMNLP}.

\bibitem[{Herbelot and Vecchi(2015{\natexlab{b}})}]{Herbelot:15}
Aur\'{e}lie Herbelot and Eva~Maria Vecchi. 2015{\natexlab{b}}.
\newblock Many speakers, many worlds.
\newblock \emph{Linguistic Issues in Language Technology}, 12(4):1--20.

\bibitem[{Johns and Jones(2012)}]{JohnsJones}
Brendan~T Johns and Michael~N Jones. 2012.
\newblock Perceptual inference through global lexical similarity.
\newblock \emph{Topics in Cognitive Science}, 4(1):103--120.

\bibitem[{Juan and Vidal(2004)}]{JuanVidal:2004}
Alfons Juan and Enrique Vidal. 2004.
\newblock Bernoulli mixture models for binary images.
\newblock In \emph{Proceedings of ICPR}.

\bibitem[{Kemp et~al.(2007)Kemp, Perfors, and Tenenbaum}]{Kemp:2007}
Charles Kemp, Amy Perfors, and Joshua~B. Tenenbaum. 2007.
\newblock Learning overhypotheses with hierarchical {Bayesian} models.
\newblock \emph{Developmental Science}, 10(3):307--321.

\bibitem[{Kipper-Schuler(2005)}]{VerbNet}
Karin Kipper-Schuler. 2005.
\newblock \emph{VerbNet: A broad-coverage, comprehensive verb lexicon}.
\newblock Ph.D. thesis, Computer and Information Science Dept., University of
  Pennsylvania, Philadelphia, PA.

\bibitem[{Kishida(2005)}]{kishida:2005:gap}
Kazuaki Kishida. 2005.
\newblock Property of average precision and its generalization: An examination
  of evaluation indicator for information retrieval experiments.
\newblock \emph{NII Technical Reports}, 2005(14):1--19.

\bibitem[{Landauer and Dumais(1997)}]{Landauer:1997tc}
Thomas Landauer and Susan Dumais. 1997.
\newblock A solution to {Plato's} problem: The latent semantic analysis theory
  of acquisition, induction, and representation of knowledge.
\newblock \emph{Psychological Review}, pages 211--240.

\bibitem[{Lazaridou et~al.(2014)Lazaridou, Bruni, and
  Baroni}]{Lazaridou:2014vt}
Angeliki Lazaridou, Elia Bruni, and Marco Baroni. 2014.
\newblock Is this a wampimuk? {C}ross-modal mapping between distributional
  semantics and the visual world.
\newblock In \emph{Proceedings of ACL}.

\bibitem[{Lazaridou et~al.(2016)Lazaridou, Marelli, and
  Baroni}]{Lazaridou:2016tb}
Angeliki Lazaridou, Marco Marelli, and Marco Baroni. 2016.
\newblock Multimodal word meaning induction from minimal exposure to natural
  text.
\newblock \emph{Cognitive Science}, pages 1--30.

\bibitem[{Maas and Kemp(2009)}]{Kemp:09}
Andrew~L. Maas and Charles Kemp. 2009.
\newblock One-shot learning with {Bayesian} networks.
\newblock In \emph{Proceedings of the 31st Annual Conference of the Cognitive
  Science Society}, Amsterdam, The Netherlands.

\bibitem[{McRae et~al.(2005)McRae, Cree, Seidenberg, and
  McNorgan}]{McRaeEtAlNorms:05}
Ken McRae, George~S. Cree, Mark~S. Seidenberg, and Chris McNorgan. 2005.
\newblock Semantic feature production norms for a large set of living and
  nonliving things.
\newblock \emph{Behavior Research Methods}, 37(4):547--559.

\bibitem[{{\'{O} S\'{e}aghdha}(2010)}]{Seaghdha:2010vi}
Diarmuid {\'{O} S\'{e}aghdha}. 2010.
\newblock Latent variable models of selectional preference.
\newblock In \emph{Proceedings of ACL}.

\bibitem[{{\'{O} S\'{e}aghdha} and Korhonen(2014)}]{OSeaghdha:Selpref}
Diarmuid {\'{O} S\'{e}aghdha} and Anna Korhonen. 2014.
\newblock Probabilistic distributional semantics with latent variable models.
\newblock \emph{Computational Linguistics}, 40(3):587--631.

\bibitem[{Ritter et~al.(2010)Ritter, Mausam, and Etzioni}]{Ritter:2010}
Alan Ritter, Mausam, and Oren Etzioni. 2010.
\newblock A {Latent} {Dirichlet} {Allocation} method for selectional
  preferences.
\newblock In \emph{Proceedings of ACL}.

\bibitem[{Roller and {Schulte im Walde}(2013)}]{Roller:13}
Stephen Roller and Sabine {Schulte im Walde}. 2013.
\newblock A multimodal lda model integrating textual, cognitive and visual
  modalities.
\newblock In \emph{Proceedings of EMNLP}.

\bibitem[{Rubinstein et~al.(2015)Rubinstein, Levi, Schwartz, and
  Rappoport}]{rubinstein2015well}
Dana Rubinstein, Effi Levi, Roy Schwartz, and Ari Rappoport. 2015.
\newblock How well do distributional models capture different types of semantic
  knowledge?
\newblock In \emph{Proceedings of ACL}, volume~2, pages 726--730.

\bibitem[{Smith et~al.(2002)Smith, Jones, Landau, Gershkoff-Stowe, and
  Samuelson}]{Smith:2002}
Linda~B. Smith, Susan~S. Jones, Barbara Landau, Lisa Gershkoff-Stowe, and
  Larissa Samuelson. 2002.
\newblock Object name learning provides on-the-job training for attention.
\newblock \emph{Psychological Science}, 13(1):13--19.

\bibitem[{Steyvers and Griffiths(2007)}]{Steyvers:07}
Mark Steyvers and Tom Griffiths. 2007.
\newblock Probabilistic topic models.
\newblock \emph{In T. Landauer, D.S. McNamara, S. Dennis, and W. Kintsch, eds.,
  Handbook of Latent Semantic Analysis}.

\bibitem[{{The BNC Consortium}(2007)}]{BNC:07}
{The BNC Consortium}. 2007.
\newblock \emph{The British National Corpus, version 3 (BNC XML Edition)}.
\newblock Oxford University Computing Services, URL:
  http://www.natcorp.ox.ac.uk/.

\bibitem[{Turney and Pantel(2010)}]{TurneyPantel:10}
Peter Turney and Patrick Pantel. 2010.
\newblock From frequency to meaning: Vector space models of semantics.
\newblock \emph{Journal of Artificial Intelligence Research}, 37:141--188.

\bibitem[{Vigliocco et~al.(2004)Vigliocco, Vinson, Lewis, and
  Garrett}]{ViglioccoEtAl:04}
Gabriella Vigliocco, David Vinson, William Lewis, and Merrill Garrett. 2004.
\newblock Representing the meanings of object and action words: The featural
  and unitary semantic space hypothesis.
\newblock \emph{Cognitive Psychology}, 48:422--488.

\bibitem[{Xu and Tenenbaum(2007)}]{XuTenenbaum:07}
Fei Xu and Joshua~B. Tenenbaum. 2007.
\newblock Word learning as {Bayesian} inference.
\newblock \emph{Psychological Review}, 114(2):245--272.

\end{thebibliography}
\bibliographystyle{ijcnlp2017}

\end{document}